\documentclass[conference]{IEEEtran}
\IEEEoverridecommandlockouts
\usepackage[noadjust]{cite}
\usepackage{amsmath,amssymb,amsfonts}
\usepackage{algorithmic}
\usepackage{graphicx}
\usepackage{textcomp}
\usepackage{xcolor}

\usepackage{booktabs}
\usepackage{acronym}
\usepackage{booktabs}
\usepackage[final]{changes}
\usepackage{todonotes}
\usepackage{siunitx}
\usepackage{makecell}
\usepackage{url}
\usepackage{multirow}
\usepackage{multicol}
\usepackage{amsthm, amssymb}

\newacro{AI}{Artificial Intelligence}
\newacro{FL}{Federated Learning}
\newacro{ML}{Machine Learning}
\newacro{NN}{Neural Network}
\newacro{CNN}{Convolutional Neural Network}
\newacro{MPU}{memory protection unit}
\newacro{TMLaaS}{TinyML-as-a-Service}
\newacro{IoT}{Internet of Things}
\newacro{CBOR}{Concise Binary Object Representation}
\newacro{CoAP}{Constrained Application Protocol}
\newacro{IID} {independent and identically distributed} 
\newacro{MCU}{Microcontroller Unit}
\newacro{I$^2$C}{Inter-Integrated Circuit}
\newacro{IETF}{Internet Engineering Task Force}
\newacro{MLP}{Multi-Layer Perceptron} 
\newacro{ANN}{Artifical Neural Networks}
\newacro{TinyML}{Tiny Machine Learning}
\newacro{AIoT}{Artificial Intelligence of Things}
\newacro{ROC}{Receiver Operating Characteristic}
\newacro{AUC}{Area Under the Curve}
\newacro{WASN}{Wireless acoustic sensor network}
\newacro{ASN}{Acoustic sensor network}
\newacro{PTQ}{Post-Training Quantization}
\newacro{TPR}{True Positive Rate}
\newacro{FPR}{False Positive Rate}
\newacro{STFT}{Short-Time Fourier Transform}
\newacro{SNR}{Signal-to-noise ratio}
\newacro{DC}{Direct Current}
\newacro{OOM}{out-of-memory}

\def\BibTeX{{\rm B\kern-.05em{\sc i\kern-.025em b}\kern-.08em
    T\kern-.1667em\lower.7ex\hbox{E}\kern-.125emX}}
\begin{document}
\bstctlcite{IEEEexample:BSTcontrol}

\title{
\footnotesize
\framebox[1.01\width]{\parbox{\dimexpr\linewidth-2\fboxsep-2\fboxrule}{If you cite this paper, please use the IEEE IS2 2024 reference: Z. Huang, A. Tousnakhoff, P. Kozyr, R. Rehausen, F. Bie\ss{}mann, R. Lachlan, C. Adjih and E. Baccelli. TinyChirp: Bird Song Recognition Using TinyML Models on Low-power Wireless Acoustic Sensors. in Proceedings of the IEEE 5th International Symposium on the Internet of Sounds. Erlangen, Germany, September 2024.}}
 \ \\ \ \\ \ \\
\huge
TinyChirp: Bird Song Recognition Using TinyML Models on Low-power Wireless Acoustic Sensors
\thanks{Corresponding author: zhaolan.huang@fu-berlin.de. }
}

\author{
    \IEEEauthorblockN{Z. Huang\IEEEauthorrefmark{1},
    A. Tousnakhoff\IEEEauthorrefmark{1},
    P. Kozyr\IEEEauthorrefmark{2},
    R. Rehausen\IEEEauthorrefmark{3},
    F. Bie\ss{}mann\IEEEauthorrefmark{2}\IEEEauthorrefmark{5}
    R. Lachlan\IEEEauthorrefmark{4},
    C. Adjih\IEEEauthorrefmark{6},
    E. Baccelli\IEEEauthorrefmark{6}\IEEEauthorrefmark{1}
    }
    \IEEEauthorblockA{\IEEEauthorrefmark{1}Freie Universit\"at Berlin
    \IEEEauthorrefmark{2}Berlin University of Applied Sciences
    }
    \IEEEauthorblockA{\IEEEauthorrefmark{3}Woks Audio GmbH
    \IEEEauthorrefmark{4}Royal Holloway University of London
    }
    \IEEEauthorblockA{\IEEEauthorrefmark{5}Einstein Center for Digital Future
    \IEEEauthorrefmark{6}Inria, France
    }

}

\maketitle
\begin{abstract}
Monitoring biodiversity at scale is challenging. Detecting and identifying species in fine grained taxonomies requires highly accurate machine learning (ML) methods. Training such models requires large high quality data sets. And deploying these models to low power devices requires novel compression techniques and model architectures. While species classification methods have profited from novel data sets and advances in ML methods, in particular neural networks, deploying these state-of-the-art models to low power devices remains difficult. Here we present a comprehensive empirical comparison of various tinyML neural network architectures and compression techniques for species classification. We focus on the example of bird song detection, and more concretely on a data set curated for studying the corn bunting bird species. We publish the data set along with all the code and experiments of this study. In our experiments we comparatively evaluate predictive performance, memory and time complexity of spectrogram-based methods and of more recent approaches operating directly on the raw audio signal. Our results demonstrate that \emph{TinyChirp} -- our approach --  can robustly detect individual bird species with precisions over 0.98 and reduce energy consumption compared to state-of-the-art, such that an autonomous recording unit lifetime on a single battery charge is extended from 2 weeks to 8 weeks, almost an entire season.  
\end{abstract}


\section{Introduction}

The typical data pipeline with bio-acoustic research requires the deployment of sensors, in which each node is battery-powered and left in the field to record environmental sound, continuously, for a long period (e.g. for a whole season). Thereafter, the recorded data is manually collected by researchers on-site, from each node, and further analyzed in laboratory. In the case of avian species for instance, only the targeted bird species is relevant within the recorded data, and the rest of the data is to be discarded ultimately. This is both a cumbersome process and a waste of resources, not only at this downstream stage in the lab, but also upstream in the field, on each node, while recording large amounts of irrelevant data. In this pipeline, continuous recording of audio thus creates a bottleneck in terms of memory and energy budgets available on individual sensors -- typically limited to a small battery, and an SD card, respectively, driven by rudimentary software running on a low-power microcontroller. Such hardware is very energy-efficient, but very limited in memory resources, with RAM memory budgets in the order of 500 kiloBytes~\cite{rfc7228}, which yields specific constraints on software embedded on such devices~\cite{hahm2015operating}.

Meanwhile, as \ac{AI} achieves excellent performance in pattern recognition of audio and biosignals, more and more bioacoustic researchers leverage \ac{ML}  related methods to improve accuracy and efficiency. 
However, until recently, such \ac{ML} models, e.g. BirdNET~\cite{kahl_birdnet_2021} were confined to lab use only, as their resource requirements (GigaBytes of RAM) are way beyond the capacity of microcontroller-based hardware available on sensors in the field. However, recent advances in TinyML, a lively field of research targeting machine learning for microcontrollers, offer a glimpse of hope that pattern recognition of bioacoustic signals might become possible, over longer periods, as required by the aforementioned use cases. 

{\bf Paper Contributions.} In this paper, we explore the possibility of using TinyML in practice, on common low-power microcontroller hardware, for a concrete use case: monitoring corn bunting birds' songs, in a rural area in the UK, over several months in a row, using a fleet of energy-efficient acoustic sensors. We aim to answer the following questions: (Q1) Can we pre-screen the audio on the low-power sensor node, so as to only store the targeted bird songs on each sensor? (Q2) How does that extend the device's lifetime, in terms of memory and energy budgets? More in detail, our contributions are as follows:

\begin{itemize}
    \item  We propose a pipeline to record, recognize and store specific bird songs on low-power microcontroller-based devices, which can be further scale-up to universal species;
    \item We developed a neural network with high accuracy on bird song recognition. Furthermore, we optimize the network with partial convolution technology to minimize the memory consumption on deployment;
    \item We propose a two-stage \added{binary} classification approach to reduce computational and storage cost and enhance the overall accuracy;
    \item We provide experimental results on both the classification performance of models and resource consumption on low-power devices;
    \item We publish a new data set of curated audio recordings snippets of corn bunting bird songs;
    \item We provide open source code\footnote{see \url{https://github.com/TinyPART/TinyChirp}} to reproduce the results on common \acp{MCU}.
\end{itemize}
\section{Background \& Related Work}


\subsection{Bird Song Recognition}
Bird song detection has seen significant recent advances using various machine learning techniques. \deleted{Recent research has explored different strategies for recognizing bird songs, with substantial progress in preprocessing audio data for recognition.}
The common prevalent approach uses deep learning methods on preprocessed audio, which transforms the signal into a spectrogram using \ac{STFT} \cite{disabato_birdsong_2021}. \deleted{Audio is thus converted into a (log) Mel spectrogram~\cite{xie_investigation_2019, xiao_amresnet_2022}. This spectrogram can subsequently be converted into a mel spectrogram \cite{xie_investigation_2019},  by applying a mel scale, followed by a log scale transformation \cite{xiao_amresnet_2022}.} This process effectively redefines the audio recognition task as an image recognition task operating on a (log) Mel spectrogram~\cite{xie_investigation_2019, xiao_amresnet_2022}. \acp{CNN} \cite{disabato_birdsong_2021, garcia-ordas_multispecies_2023, xie_investigation_2019} and Residual Neural Networks (ResNets) \cite{xiao_amresnet_2022, hu_deep_2023} are widely utilized for such image recognition tasks in bird song detection. \deleted{For instance, one study employed CNNs to recognize the songs of 17 bird species \cite{garcia-ordas_multispecies_2023}, while another study utilized ResNets to classify the songs of 19 bird species \cite{xiao_amresnet_2022}.} BirdNET \cite{kahl_birdnet_2021}, a well-known neural network for bird song recognition, is derived from the family of residual networks and can recognize 6,000 of the world’s most common species at the time of writing.

Another approach involves transforming the audio signal into Mel-Frequency Cepstral Coefficients (MFCCs), which are then used as input for bird song classification \cite{zhang_novel_2023, stowell_automatic_2019, garcia-ordas_multispecies_2023}. Alternatively, feature extraction using wavelet decomposition can be employed for this purpose \cite{selin_wavelets_2006}. Additionally, certain studies combine different architectures \added{\cite{zhang_novel_2023,xie_investigation_2019}.} \deleted{; for example, \ac{CNN} and Transformer \cite{zhang_novel_2023}, and \ac{CNN} and VGG \cite{xie_investigation_2019} to enhance performance.} While these methods are effective for bird song recognition, their computational requirements are typically too demanding for execution on microcontrollers.

 \added{In this work we focus on bird songs, rather than calls. The main reasons for this are the higher complexity of bird songs, their role in sexual selection, and the generally richer behavioral information they provide. 
 }

\subsection{Bioacoustic Audio Datasets}
High-quality and diverse datasets are crucial for training models to accurately recognize birds and differentiate between species by their unique songs. The Xeno-canto database\footnote{see \url{https://www.xeno-canto.org}} is one of the largest and most comprehensive collections of bird sounds, containing over 500,000 recordings from more than 10,000 species. \deleted{In \cite{garcia-ordas_multispecies_2023}, 2,699 bird song audio recordings from the Xeno-canto database, ranging in length from 20 to 0.74 seconds, were used to classify 17 bird species.}
Another extensive collection of bird audio recordings is available in the Macaulay Library\footnote{see \url{https://www.macaulaylibrary.org/}}, renowned for its high-quality bird audio recordings and detailed metadata\deleted{, which includes information about species, location, and context}. The combination of the Macaulay Library and Xeno-canto audio recordings was used to train BirdNet \cite{kahl_birdnet_2021}.
In \cite{xie_investigation_2019}, the public dataset CLO-43DS contains recordings of 43 different North American wood-warblers. Another notable collection is the Birdsdata dataset with \deleted{audio files of} 20 bird species from the Beijing Academy of Artificial Intelligence (BAAI) repository\deleted{, which is frequently used in research and model training} \cite{zhang_novel_2023, hu_deep_2023, xiao_amresnet_2022}.
\added{The BEANS benchmark \cite{hagiwara2022beansbenchmarkanimalsounds}, designed to evaluate \ac{ML} algorithms for bioacoustics tasks across various species, includes 12 public datasets on birds, mammals, anurans, and insects.}
To complement datasets focused on bird species, the Google AudioSet \cite{kahl_birdnet_2021} and Urbansound8K \cite{hu_deep_2023} datasets are often used to include non-bird species sounds.

\subsection{\ac{TinyML}}
For models to operate on low-power devices, they must be compact and computationally efficient. Studies have demonstrated the use of lightweight CNNs for \added{various Internet of Sounds applications \cite{maayah_limitaccess_2023, atanane_smart_2023,donati_tiny_2023,fang_fall_2021,hussain_swishnet_2018, zhu-zhou_computationally_2023}.} \deleted{speech recognition and age classification \cite{maayah_limitaccess_2023}, water leakage detection \cite{atanane_smart_2023}, acoustic defect localization \cite{donati_tiny_2023}, fall detection for the elderly \cite{fang_fall_2021} and other tasks \cite{hussain_swishnet_2018, zhu-zhou_computationally_2023}.} Other model architectures, such as tiny vision transformers have also been employed for \added{audio} classification tasks in various studies \cite{jinyang_yu_tiny_2023, busia_tiny_2024, liang_mcuformer_2023, yao_cnn-transformer_2023, wyatt_environmental_2021}. Moreover, various quantization techniques are applied to models, aiming to reduce their size to fit within the constraints of low-power devices \cite{xiao_smoothquant_2022, park_urban_2023, choi_reducing_2022}.

Creating spectrograms from audio signals can be power-intensive; hence, some studies use instead raw time-series data as input for neural networks \cite{abdoli_end--end_2019, sawhney_latent_2018, shalini_mukhopadhyay_time_2024}. \deleted{For instance, single-channel EEG was used as input for a model based on CNN and Transformer for sleep stage classification \cite{yao_cnn-transformer_2023}.} Notably, raw audio signals have been used for urban sound analysis \cite{sang_convolutional_2018, park_urban_2023}. The advantages of using time-series data include reduced computational load and suitability for TinyML on low-power devices.

\added{In a nutshell: the integration of \ac{TinyML} and acoustic sensors has become a hot topic in the field of Internet of Sounds~\cite{turchet2023internet}. As demonstrated in prior bioacoustic applications such as \cite{schulthess2023tinybird, disabato_birdsong_2021, miquel2023energy}, TinyML promises to offer new, appealing combinations of high accuracy and resource-efficiency.}

\subsection{Embedded Software Platforms for TinyML} The widely used model transpiler TVM (Tensor Virtual Machine~\cite{chen2018tvm}) has recently been extended with uTVM, providing automated transpilation and compilation for models output by major ML frameworks (TFLM, Pytorch, etc.). As such uTVM exposes low-level routines and  optimizes these for execution on different processing units, including for targets such as a large variety of microcontrollers. Prior works such as~\cite{banbury2020benchmarking}, \cite{poloTinyBenchmarksEvaluatingLLMs2024} or MLPerfTiny~\cite{banbury2021mlperf} focused on the production, performance and analysis of standard benchmark suites of representative TinyML tasks on different microcontrollers. Conversely, prior work such as U-TOE~\cite{huang2023u-toe} or RIOT-ML~\cite{huang2024riot-ml} provide embedded operating system integration of TinyML, facilitating TinyML benchmarking and continuous deployment over low-power wireless network links such as IEEE 802.15.4 or BLE.

\section{Bioacoustic Monitoring Scenario}

\begin{figure}
    \centering
    \includegraphics[width=0.8\linewidth]{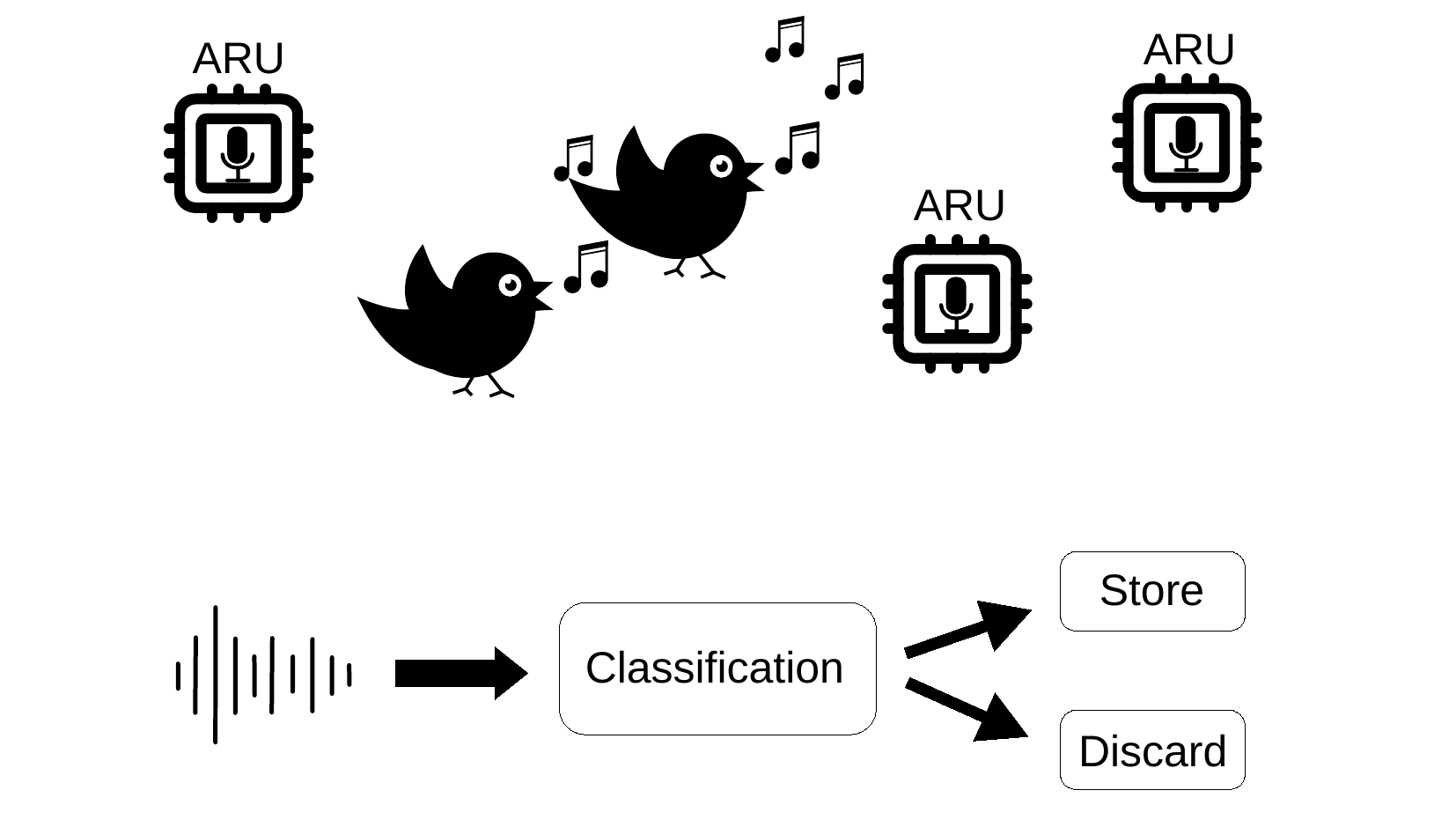}
    \caption{\textit{Top}: Low-power autonomous recording units (ARU) distributed spatially to monitor birds. \textit{Bottom}: audio sample classification on an ARU.}
    \label{fig:scenario_bird_song_recognition}
\end{figure}

As depicted in Figure~\ref{fig:scenario_bird_song_recognition}, a network of battery-powered, autonomous recording units (ARUs, i.e. microcontroller-based acoustic sensors) is deployed across a monitored area. An example of ARU is given in~\cite{zandberg2023songbeam}.

These sensors remain in the field for an entire season, typically around six months. Locations are often remote and hard to access and devices may be distributed over a relatively wide area, making visiting them time-consuming. Birds typically do not sing consistently during the whole day, while each bird has several different song types that need to be regularly sampled. Moreover, each individual bird moves around its individual territory, singing from a range of song posts, not all of which will be adequately recorded from any one location. 

On the network aspect, the sensors are distributed approximately evenly, ensuring that each sensor can be wireless connected by more than two neighboring sensors. This arrangement facilitates approximate triangulation through distance estimation based on signal strength. The placement process can be streamlined using LEDs and a basic ultra-low-power wireless protocol, such as IEEE 802.15.4 or LoRa. For space reason, we do not detail \deleted{further} network aspects in the paper, but rather focus on the on-device machine learning aspects. \deleted{We nevertheless keep an eye on the total memory footprint on the device, to ensure that the operating system including application TinyML code, the OS and a typical wireless low-power network stack fit in typical resource budgets on common microcontroller-based boards.}

Although our work is designed around one very specific real-world research scenario (monitoring common Corn Bunting bird songs) the architectures should be readily applicable to other audio data classification tasks. \deleted{Examples of related scenarios to which our method could be extended include, surveying remote areas for targeted endangered species of conservation value, or broader surveys of songbird biodiversity.  Similar devices would even be useful in the search for very elusive species such as Ivory-billed Woodpecker~\cite{collins2021woodpecker}. }

\subsection{Corn Bunting Monitoring Use-Case}
\label{subsec:corn_bunt_use_case}

SongBeam~\cite{zandberg2023songbeam} microcontroller-based recorders have been used to monitor corn bunting (\textit{Emberiza calandra}) birds in Oxfordshire, UK. 
A set of 30-40 devices have been deployed since 2022, currently comprising 3 complete breeding seasons (February - July).

SongBeam devices are based on an ARM Cortex-M microcontroller, run on 4 D-cell batteries and record 4-channel WAV files onto 128 \si{\giga\byte} microSD cards. To maintain such a fleet of sensors, devices are currently checked approximately every 2 weeks, especially because memory space is soon exhausted on the microSD cards. 

The above deployment has been used to produce a significant part of the dataset we present in Section \ref{subsec:dataset}.
Our experience using SongBeam and a preliminary analysis of the raw dataset we produced shows that less than 10\% of recording time contains useful recordings. 
Given the extended deployment period (6 months), remaining low-power while improving storage efficiency is thus paramount. 
\deleted{Moreover, current prototypes use solar power, further emphasizing that the bottleneck is downloading data from the microSD cards to free space on-board.} 

This description of the limitations of using SongBeam recorders to monitor corn buntings is likely to apply to many biomonitoring scenarios: solar power may allow for extending deployment lifetime in remote locations, but storage of recordings is a limiting factor.


\section{Evaluation Metrics}
\label{sec:eval_metr}

On the one hand, as the basic part of bird song recording, the model should identify the target bird song as much as possible, potentially requiring a complex structure with a large size. On the other hand, the limited resource budget allows only simple models to be deployed on \deleted{low-power} \ac{IoT} devices.

Thus we considered two orthogonal types of metrics to evaluate the prediction performance on bird song classification and resource usage on low-power \ac{IoT} devices of TinyML models. Preliminary results are presented in Section \ref{sec:perf_eval}.

\subsection{On-device Resource Usage}
We performed extensive profiling of all models on low-power devices to capture their effective memory and time complexity in a real-world application scenario. The metrics are helpful to investigate the energy efficiency of the models, complementing the commonly used metrics for predictive performance. 

{\bf Memory (RAM) Consumption --}
This metric measures the amount of dynamic memory space (primary RAM) consumed by the model during inference. It reflects the memory footprint of the model activation and is important for low-power devices that have limited memory resources. \deleted{Efficient memory utilization allows for the deployment of larger and more complex models on such devices.}

{\bf Storage (Flash memory) Consumption --}
This metric quantifies the amount of storage space, typically in terms of Flash memory region, required to store the compute instruction and associated parameters. 
\deleted{Minimizing storage consumption allows for accommodating multiple models on the device or orchestrating with other essential applications.}

{\bf Computational Latency --}
This metric measures the time consumption of performing inference for each input sample at the model level. It reflects the inference speed of the model on the low-power device and plays a crucial role in real-time or latency-sensitive applications. \deleted{Core clock frequency, cache strategies and communication latency between memory and working core have a great impact on this indicator.}

{\bf Energy Consumption --} 
This metric is crucial for battery-operated and resource-constrained devices, where efficient energy usage can significantly impact device longevity and performance. Energy consumption encompasses both active power (when the device is triggered to perform tasks) and idle power (when the device is in sleep mode).

\subsection{On-device Prediction Performance}

{\bf Accuracy --}
It is defined as the ratio of correctly predicted instances to the total instances in the dataset. While accuracy is intuitive, it can be misleading, especially in highly imbalanced datasets. 
To address this limitation, accuracy should be combined with other metrics that offer a more detailed view of model performance.

{\bf Precision and Recall --}
They are two fundamental metrics of classification models used to evaluate the ability to distinguish true positives (TP) from false positives (FP) and false negatives (FN), particularly in the context of imbalanced datasets. 

{\bf F-Score --}
There is often a trade-off between precision and recall; increasing one can lead to a decrease in the other. 
As two common instances, $F_1$-score weighs them evenly, while $F_2$-score treats recall as two times more important than precision, applying in scenarios where false positives are more tolerant than false negatives.

{\bf \ac{ROC} Curve --}
While the above metrics are sensitive to class imbalance -- which occurs in many species classification and detection tasks -- there are other metrics that are more robust towards class imbalance. The receiver-operator characteristic (\ac{ROC}) curve depicts the  \ac{TPR} and \ac{FPR} across different decision thresholds.
\deleted{The \ac{AUC} is a key metric derived from the ROC curve that quantifies the overall performance of the model and is invariant with respect to class imbalance. An \ac{AUC} of 1.0 indicates perfect classification, while an \ac{AUC} of 0.5 suggests no discriminative ability. } 

\section{Methodology}
We used a combination of project-specific and publicly accessible data to train and validate our classification models. In the data pre-processing phase, the raw audio signals were further segmented, labeled, down-sampled and divided into different groups to generate classification datasets. The corresponding Mel-spectrograms of the pre-processed audio segments were also created for spectral-based methods (models). During the pre-processing stage, we conducted a pilot analysis to establish guidelines and determine specific hyper-parameters for downsampling and spectrogram generation.

\subsection{Data Acquisition}
\label{subsec:dataset}
The publicly accessible data originated from Macaulay, Xeno-Canto, Google AudioSet, while the project-specific data were previously collected in a long-term research of bird song patterns. \added{In order to ensure generalization across a wide variety of conditions for the target species, the corn bunting, we gathered as many recordings as possible from a heterogeneous suite of publicly available repositories, in addition to the custom data set collected. We aimed at increasing precision of all models by including a wide variety of non-target sounds. Our data set is made publicly available}\footnote{see \url{https://github.com/TinyPART/TinyChirp/tree/main/datasets}}.

\begin{itemize}
    \item Oxfordshire Corn Buntings. This project-specific library contains recordings of corn buntings along a transect of approximately 20~\si{\kilo\meter} in Southern England. Corn buntings sing in a mosaic-like pattern of geographical variation called dialects; our sample contains approximately 6 different dialects. The recordings were performed with directional parabolic microphones as described in~\cite{zandberg2023songbeam}.  
    
    \item Macaulay Library. This library contains the necessary audio recordings for our target species as well as the other identified species. There are a total of 278 entries of audio recordings for the target species. \deleted{We filtered the audio files by the song tag and removed 16 entries due to unclear metadata and missing catalog numbers.} For other species as non-target, we limit the selection to a maximum of 30 recordings per species and choose those with the highest average community rating, acquiring a total of 1468 recordings.
    
    \item Xeno-Canto. This library also contains recordings of target species and other species. We retrieved the recordings only marked with \emph{song} and rated with the best quality level, resulting in 303 recordings of target species and 9622 recordings of other species as non-target.
    
    \item Google AudioSet. This dataset provides environmental sounds which contain non-bird sounds. \added{We create a list of excluded categories to avoid overlap with other bird sounds. This data set includes weather-related sounds, such as rain, wind, thunderstorms, insects, other animal and human-related sounds, such as church bells, trucks, etc.}
\end{itemize}

After gathering all audio recordings from the above libraries, we used BirdNET \added{with the confidence threshold at 0.92} to identify and chop corn-bunting segments from all datasets, and labeled them as target bird songs; all segments were truncated or zero-padded to the length of 3 seconds. \added{This high confidence threshold is intended to minimize incorrect labels, although it does not entirely eliminate the label noise, which could potentially skew the performance metrics of our models, particularly in cases where the models are trained on mislabeled data.} Meanwhile, we chose other 3-s segments randomly and ensured no overlap with corn-bunting segments, with labeled as non-target bird songs. \added{To avoid data leakage, we implement an additional function that segments occurring sequentially in the source audio file were not distributed across the training, validation and test sets.} Table~\ref{tab:data_dist_datasets} shows the distribution of target and non-target segments over all datasets.

\subsection{Data Pre-processing}
\label{subsec:preprocessing}

This phase contains the following steps:

\begin{enumerate}
    \item We divided the segments into training, validation and test sets with the ratio of $80:10:10$.

    \item All segments were downsampled to 16 \si{\kilo\hertz} using zero-order holder. \added{These downsampled segments constituted the audio (waveform) dataset.}
    
    \item We transformed the downsampled audio segments into Mel-Spectrograms, and grouped them with the same splitting ratio to form the spectral dataset.
\end{enumerate}

\begin{figure}
    \centering
    \includegraphics[width=0.8\linewidth]{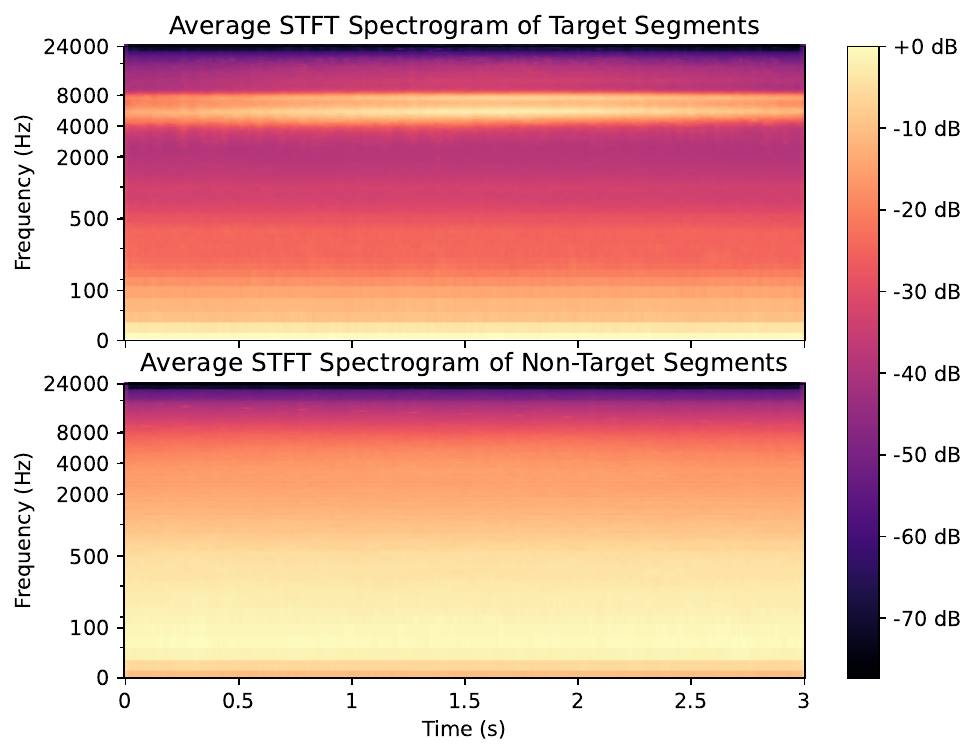}
    \caption{Average \ac{STFT} of target and non-target audio segments.}
    \label{fig:avg_stft_bird_vs_non_bird}
\end{figure}

{\bf Pilot Analysis --}
We averaged the \ac{STFT} spectrograms of target and non-target segments in the training dataset to investigate their frequency characteristics, as depicted in Figure~\ref{fig:avg_stft_bird_vs_non_bird}. Obviously, a bright band lays on the spectrogram of target segments roughly between 4000 and 8000 \si{\hertz}, hints that a sample rate with 16 \si{\kilo\hertz} should be sufficient to preserve all frequency components of corn-bunting song according to Nyquist–Shannon theorem. A highpass filter can also be applied to eliminate low-frequency noise without damaging target songs. Thus, we downsampled the segments from 48 \si{\kilo\hertz} to 16 \si{\kilo\hertz} to reduce the resource consumption and improve the efficiency of classification phase. In the baseline model we also designed a highpass filter to enhance \ac{SNR}.

{\bf Mel-Spectrogram --}
Mel-spectrogram is commonly used in audio classification tasks due to its ability to closely approximate human auditory perception. To create log Mel-spectrograms, we first generated magnitude spectrograms of \ac{STFT} for all downsampled audio segments. The \ac{STFT} was performed with a Hann window with a width of 1024 samples and a step size of 256 samples. This process generated 184 window frames and 513 frequency bins. Thereafter, the magnitude spectrograms were mapped onto the Mel scales with 80 Mel bins. In creating the Mel-spectrograms, we employed a sampling rate of 16 \si{\kilo\hertz} and focused on the frequency range of 80-8000 \si{\hertz}. \deleted{This frequency range was chosen because it encompasses the typical frequency range of most bird songs, including that of the target Corn Bunting.} \added{This also allows \ac{ML} models to capture subtle patterns outside the primary band (4000 \si{\hertz} to 8000 \si{\hertz}), adapt to variations, and improve overall classification accuracy by leveraging information that a simple filter might overlook.} Lastly, we converted the Mel-spectrograms into a logarithmic magnitude scale to obtain log Mel-spectrograms, with the shape of $184 \times 80$. 

\begin{table}[]
    \centering
    \caption{Data distribution over different Datasets.}
    \begin{tabular}{lll}
    \toprule
    Dataset & \# Target  & \# Non-Target \\\midrule
    Oxfordshire Corn Buntings & 1566 & 0\\
    Macaulay Library & 1059 & 2057\\
    Xeno-Canto & 2111 & 2621\\
    Google AudioSet & 0 & 4651\\ \midrule
    Summary & 4736 & 9329 \\
    \bottomrule
    \end{tabular}
    \label{tab:data_dist_datasets}
\end{table}

\section{Baseline \& Decision Strategy}

We propose an approach to optimizing the screening process by combining a high-speed signal processing step (hereafter named "baseline") with the precision of an \ac{TinyML} model, ensuring reliable results while maintaining resource-consumption efficiency.

\subsection{Baseline}
\label{subsec:baseline}
We developed a lightweight pre-selector signal processing step, not relying on machine learning. First, the audio segments are normalized by min-max schema. Thereafter, a 9th-order highpass Butterworth filter with the cut-off frequency of 7000~\si{\hertz}\footnote{\added{The parameters of the filter were determined by hyper-parameter search for optimized $F_2$-score on training dataset.}} is imposed on the normalized segments, in order to suppress noise and non-relative sound in low-frequency and to preserve only target components. Finally, we calculate the signal power $P$ of the filtered segments,

\begin{equation}
    P = \frac{1}{N} \sum_{n=1}^{N} x(n)^2,
\end{equation}
where $x(n)$ and $N$ denote the data points and the length of a segment, respectively. 
We define two thresholds: $t_{low}$ and $t_{high}$. If $P$ is smaller than $t_{low}$, the \ac{MCU} remains in the idle/low-power state, and the audio sample is discarded directly. Else, $P$ is compared to $t_{high}$. Depending on this result, and on the decision strategy (see below), the audio sample is either stored or further analyzed via inference using the machine learning model. \added{Both $t_{low}$ and $t_{high}$ were tuned on training dataset for optimized recall and $F_2$-score, respectively. The evaluation results are presented in Table~\ref{tab:mdl_threshold_opt_f2}.}


\subsection{TinyChirp Decision Strategy}
\label{subsec:decision_strategy}


We employ a double-phase approach for enhanced accuracy and efficiency of bird song recognition, combining the baseline with a \ac{TinyML} model for further verification. The process is as follows:
\begin{enumerate}
    \item {\bf Step 1:} \emph{Baseline Filtering --} Pre-screening is conducted using the baseline pre-processing step described in Section \ref{subsec:baseline}, designed to provide a quick and efficient preliminary analysis. If $P$ is smaller than $t_{low}$, the microcontroller remains in the idle/low-power state, the audio sample is discarded and the process aborts. Else proceed to Step 2.
    


    \item  {\bf Step 2:} \emph{Inference-based Classification --} If the power-saving flag is set and if $P$ is smaller than $t_{high}$ a \replaced{\ac{TinyML}-based classification with higher accuracy is conducted via model inference.}{higher-accuracy classification is conducted via inference using \ac{TinyML} model.} If the audio sample is not classified as the target, the audio sample is discarded and the process aborts. Else proceed to Step 3.

\item {\bf Step 3:} \emph{Storage --} Without further processing, the audio sample is logged, i.e. stored on the SD card. 

\end{enumerate}

Note that the decision strategy can therefore be configured with two "knobs". On one hand the power-saving flag determines skipping (or not) some of the computation and can thus decrease energy consumption. On the other hand, different TinyML models can be inserted in Step 2.


Next, we thus study two families of TinyML models for Step 2: the first category of model takes a Mel-Spectrogram as input, while the second category of model takes time-series as input. Table~\ref{tab:struct_tinyml_models} presents a summary of the explored \ac{TinyML} model structures.

\section{Classification Based on Spectrograms}

In contrast to the computer vision community, where the transition from frequency decomposition based feature extraction to neural feature extraction on raw image data introduced in 2012 led to a significant increase in predictive performance \cite{krizhevsky2012imagenet}, neural feature extraction on raw audio time series did not contribute to a comparable breakthrough yet. Instead most ML techniques for \emph{audio} pattern recognition (somewhat surprisingly) were relying on \emph{image} processing pipelines\deleted{ -- sometimes with neural networks trained on ImageNet data}. Basically, image classification is performed, using a neural network, analyzing the spectrogram rendering of the audio trace.

\added{Unlike the spatial dimensions in image classification, the axes of the input represent time and frequency in audio classification. Moreover, the pattern structure differs significantly: natural objects in images often have well-defined, closed contours, while spectrograms of sounds tend to be wide-band and sparsely distributed across the frequency-time domain. These distinctions require tailored approaches to effectively model and interpret the data in audio classification tasks.}

For this reason, we initially consider the below two models, based on state-of-the-art neural network architectures.

{\bf CNN-Mel --}
This model contains two 2D convolutional layers for Mel-spectral inputs as feature extraction and two fully connected layers as classifier. \deleted{Both convolutional layers utilize 4 filters with a $3 \times 3$ kernel size and ReLU activation function, followed by $2 \times 2$ max pooling to reduce spatial dimensions.} The output of the classifier are normalized by Softmax function as well.

{\bf SqueezeNet-Mel --} 
This model is based on SqueezeNet \cite{iandola2016squeezenet}, an advanced backbone focused on optimizing the efficiency of computer vision applications by strategically reducing parameters, with comparable performance to AlexNet \cite{krizhevsky2012imagenet}. It leverages a so-called \emph{Fire module} to achieve higher efficiency compared to standard convolutions with only a slight decrease in accuracy. \deleted{The \emph{Fire module} significantly reduces the number of parameters by the squeeze layer ($1\times1$ convolutions), which exceedingly reduces the number of channels before the more parameter-heavy $3\times3$ convolutions in the expand layer. This reduction in intermediate channels means fewer parameters overall.}
We aligned its input shape with the Mel-spectrogram and tailored the output for binary classification. 
\deleted{Also offering potential for further compression using quantization techniques, SqueezeNet is thus appealing given memory constraints.}

\textbf{Limitations of Spectrograms on Microcontrollers}
Practical experience on low-power microcontroller-based devices has shown however that producing and manipulating mel-spectrograms from audio signal streams on such devices is problematic. It inserts an additional step in the processing pipeline, which increases latency and memory requirements. Prior work (such as~\cite{nordby2019environmental}, section 6.2) concludes that the CPU bottleneck is not just the model inference time, but also the spectrogram calculation, which, compared to inference alone, almost doubles latency. Other previous work such as~\cite{wang2022keyword} measures on a quite powerful STM32F7 microcontroller that computing and writing in memory the mel-spectrogram of 30 columns and 40 frequency bands takes approximately 1 second.  Note that these spectrogram dimensions are much smaller than our requirements (184 columns × 80 frequency bands) hence even more latency can be expected in our case.
For these reasons, we next explore pipelines which skip the spectral pre-processing stage as described below.

\begin{table}[]
    \centering
    \scriptsize
    \caption{TinyML Models: Structure and Characteristics.}
    \begin{tabular}{llll}
    \toprule
    Model &     Layer  & Input Shape & Output Shape \\\midrule

    \makecell[l]{CNN-Mel\\(25.6K parameters)}  & $3\times3$ Conv2D + ReLU  & $184 \times 80 \times 1$  &  $182\times78 \times4$    \\
                & MaxPooling  & $182\times78 \times4$ &  $91\times39 \times4$    \\
                & $3\times3$ Conv2D + ReLU  & $91\times39 \times4$ &  $89\times37 \times4$    \\
                & MaxPooling  & $89\times37 \times4$ &  $44\times18 \times4$    \\
                & Reshape  & $44\times18 \times4$ &  $3168\times1$    \\
                & FC  + ReLU  & $3168\times1$ &  $8\times1$    \\
                & FC  + Softmax & $8\times1$ &  $2\times1$    \\ 
                \midrule
    \makecell[l]{SqueezeNet-Mel\\(727K parameters)} & \makecell[l]{Same as SqueezeNet \cite{iandola2016squeezenet};\\Input and output layer are\\tailored to fit the data.} & $184 \times 80 \times 1$ & $2 \times 1$ \\

    \specialrule{1.5pt}{2pt}{2pt}

    \makecell[l]{CNN-Time\\(748 parameters)}  & $3 \times 1$ Conv1D + ReLU  & $1 \times 48000$  &  $4 \times 48000$    \\
                & MaxPooling & $4 \times 48000$ & $4 \times 24000$ \\
                & $3 \times 1$ Conv1D  & $4 \times 24000$  &  $8 \times 24000$    \\
                & Dropout 0.25 & $8 \times 12000$ & $8 \times 12000$ \\
                & Average Pooling & $8 \times 12000$ & $8 \times 1$ \\
                & FC + ReLU & $8 \times 1$ & $64 \times 1$ \\
                & FC + Softmax & $64 \times 1$ & $2 \times 1$ \\ 
                \midrule

    \makecell[l]{Transformer-Time\\(1.6K parameters)} &  Conv1D + ReLU & $1 \times 48000$ & $16 \times 48000$\\
                    & MaxPooling & $16 \times 48000$ & $16 \times 24000$\\
                    & Dropout 0.25 & $16 \times 48000$ & $16 \times 24000$ \\
                   & Average Pooling & $16 \times 24000$ & $16 \times 1$\\
                    & SingleHeadTransformer & $16 \times 1$ & $16 \times 1$\\
                    & FC + Softmax & $16 \times 1$ & $2 \times 1$\\
    
    \midrule
    \makecell[l]{SqueezeNet-Time\\(31.1K parameters)} & \makecell[l]{SqueezeNet as backbone;\\Conv2D $\rightarrow$ Conv1D;\\reduce filter number \\by 70\%.} & $1 \times 48000$ & $2 \times 1$ \\

    \bottomrule
    \end{tabular}
    \label{tab:struct_tinyml_models}

\end{table}

\begin{table}[]
    \centering
    \caption{Lightweight variant of the Fire module of 1D SqueezeNet. The number of filters in original SqueezeNet is represented by $x$. }
    \begin{tabular}{llll}
    \toprule
    
    Name &     Layer  & Number of filters  \\\midrule
    Squeeze $1 \times 1$    & Conv1D &  \(3\times \lfloor 0.3\times x \rfloor\)    \\
    Expand $1 \times 1$  & Conv1D &  \(4\times \lfloor 0.3\times x \rfloor\)    \\
    Expand $3 \times 1$  & Conv1D &  \(4\times \lfloor 0.3\times x \rfloor\)     \\

    \bottomrule
    \end{tabular}
    \label{tab:struct_lightweight_fire_module}
\end{table}

\section{Classification Based on Time-series}

Contrary to the models described in the previous section and inspired by the success of neural feature extraction in the computer vision domain \cite{krizhevsky2012imagenet}, the pipelines we aim at next take directly the raw audio signal time-series as input. More precisely, we designed the three models described below. \added{We inserted the dropouts before average pooling to gain a more stable output~\cite{kim2023use}.}

{\bf CNN-Time --}
This simple model performs feature extraction with two sequential (1D) temporal convolutional layers followed by average pooling. A max pooling is inserted in-between to reduce the dimension of feature maps and enhance the non-linearity of the network. Two fully connected layers act as classifier at the end of the network, with the probability outputs normalized by Softmax function.  

{\bf Transformer-Time --}
Inspired by \cite{elliott2021tinytrans}, we designed an efficient, attention-based model with only one temporal convolutional layer and one single-head transformer. The convolutional and pooling layers serve as feature extraction to transfer raw audio signal into embeddings for the transformer. The activations of the fully connected layer are normalized by Softmax function as well.


{\bf SqueezeNet-Time --}
This model is based on SqueezeNet \cite{iandola2016squeezenet}. We tailored this basis to align the input shape with the audio segment, and narrowed the output channels for binary classification. To adapt the backbone to time-series input, all 2D convolutional layers are replaced by temporal convolutional layers. Furthermore, to have a more compact structure, we decreased the filter number of all convolutional layers by roughly 70\%, as presented in Table \ref{tab:struct_lightweight_fire_module}.

\section{Additional Model Optimizations for TinyML}
\label{subsec:opt_strategies}
In this study, we used two optimization techniques to compress the models and further reduce their memory consumption: model quantization on the one hand and on the other hand partial convolution, as described below.



\subsection{Model Quantization}
Quantization reduces the model size and inference time by converting the weights and activations from higher precision (e.g., 32-bit floating-point) to lower precision (e.g., 8-bit integer) \cite{nagel2021white}. In this study we adopted \ac{PTQ} with weights and activations quantized in 8-bit integer. To avoid substantial loss of prediction performance, the training dataset was used to find the optimal scale factor and zero-point of the activations \cite{jacob2018quantization}.

\subsection{Partial Convolution}
We observed that the outputs of the Conv1D layer require significant memory, which impedes deployment on resource-constrained tiny devices. As presented in Table~\ref{tab:struct_tinyml_models}, the peak memory consumption occurs at the first Conv1D layer both in CNN-Time and Transformer-Time with 768 \si{\kilo\byte} and 3 \si{\mega\byte}, respectively. This indicates that deployment on resource-constrained tiny devices is impractical.

To address this issue and inspired by \cite{pinckaers2020streaming}, 
we exploited the fact that average pooling can be computed iteratively point-by-point on the output channels of the final Conv1D layer over a small sliding window of inputs. Each point in the output channels depends only on a small subset (kernel window) of the outputs from the previous layer. That is, for a block of $L$ Conv1D layers following by an average pooling layer, its output $y$ can be iteratively computed on the input sequence $x(n), n= 1 \dots N$ as following:
\begin{align}
    y_j(k) &= y_j(k-1) + \frac{1}{N_L} A^L_j(k), \\
    k &= 1 \dots N_L, j = 1 \dots C_L . \notag\\ 
    A^l_c(n) &= \sum_{i=1}^{C_{l-1}} W^l_c(i) \cdot [ A^{l-1}_i(n-\frac{K^l}{2}) \cdots \\ 
                                            &\phantom{{}=4}A^{l-1}_i(n) \cdots A^{l-1}_i(n+\frac{K^l}{2}) ]^T \notag \\
    A^0(n) &= x(n) , \\
    l &= 1 \dots L, c = 1 \dots C_l, n = 1 \dots N_l, N_0 = N \notag
\end{align}
where $y_j(N_L)$ is the result of the average pooling of the $j$-th channel; $A^l_c(n)$ denotes the $n$-th point in channel $c$ of the $l$-th Conv1D layer calculated with kernel weights $W$. Each Conv1D layer contains $C_l$ channels with filter size of $K^l$ and output size of $N_l$. Without loss of generality, the formula of partial convolution contains only the linear components and considers only one-dimension case for the sake of simplicity; it can be generalized in high-dimension and integrated with non-linear building blocks (e.g., stride, non-linear activations, pooing layers, etc.).

\begin{figure}
    \centering
    \includegraphics[width=0.8\linewidth]{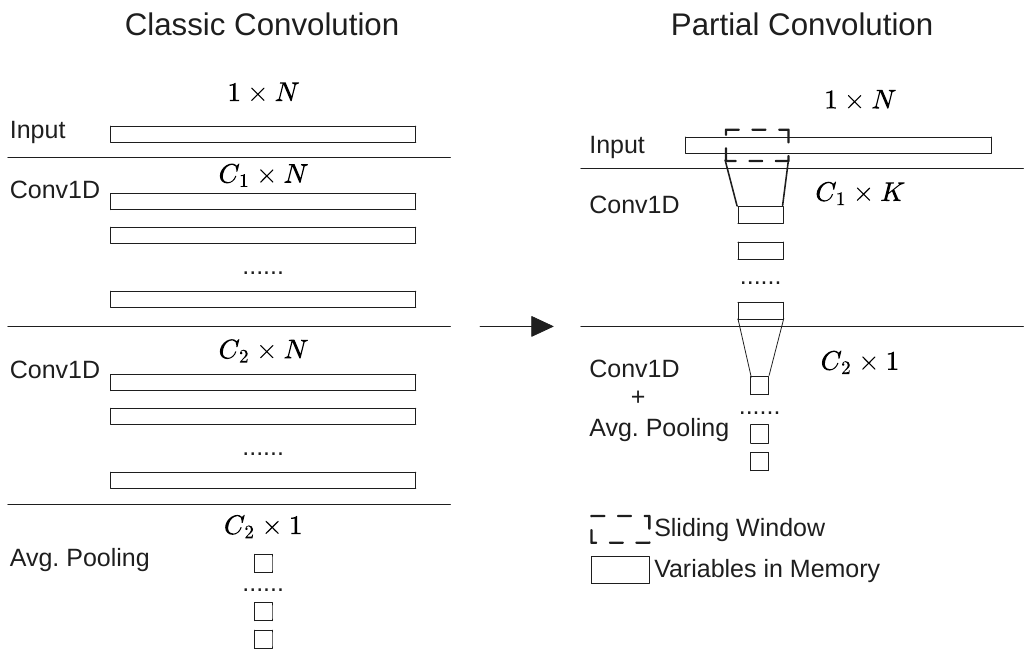}
    \caption{Partial convolution reduces memory consumption by applying small sliding window on inputs.}
    \label{fig:partial_convolution}
\end{figure}

Figure~\ref{fig:partial_convolution} provides a comparison between classic convolution and partial convolution. Unlike classic convolution where entire channels ($C \times N$) are computed and stored before being processed by the next layer, partial convolution requires only a small part of the channels ($C \times K, K << N$) for each layer. In our case with $K=3$ and $N=48000$, it can be expected roughly $16000 \times$ smaller memory consumption, making the implementation more suitable for tiny devices. We applied this strategy to the Conv1D layers combined with average pooling in the CNN-Time and Transformer-Time architectures.




\section{TinyChirp Performance Evaluation}
\label{sec:perf_eval}
In the following, we conducted a comprehensive evaluation of the discriminative performance of TinyChirp with different TinyML models, as well as of resource consumption and computation time on typical low-power boards based on microcontrollers.

\subsection{Classification Performance Comparison}
\label{sec:pred_perf_models}



\begin{figure}
    \centering
    \includegraphics[width=0.8\linewidth]{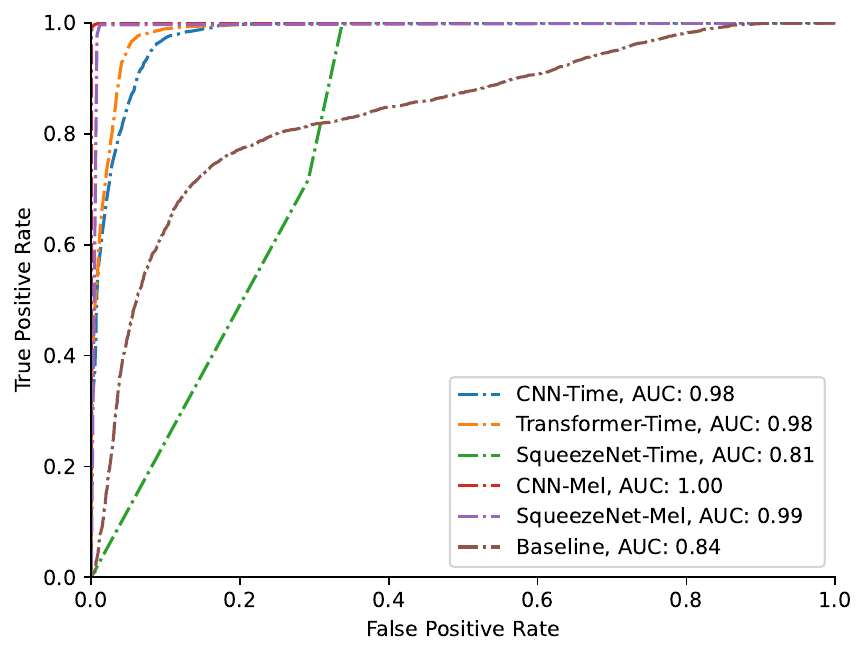}
    \caption{\ac{ROC} curves of \ac{TinyML} models and the corresponding \acp{AUC}, evaluating on training dataset.}
    \label{fig:model_roc_auc}
\end{figure}

Our evaluation process began with the generation of \ac{ROC} curves and the corresponding \acp{AUC} on the training dataset, providing a clear picture of the models' abilities to distinguish between classes. As shown in Figure~\ref{fig:model_roc_auc}, the spectrogram-based models -- CNN-Mel and SqueezeNet-Mel -- achieved the best classification performance with \ac{AUC}s of 1.0 and 0.99, respectively, followed by time-series models -- CNN-Time and Transformer-Time -- both with 0.98 \ac{AUC}. Unexpectedly, with a more complex structure, SqueezeNet-Time performs worst (lower \ac{AUC} than the baseline).

\begin{figure*}
    \centering
     \includegraphics[width=0.8\linewidth]{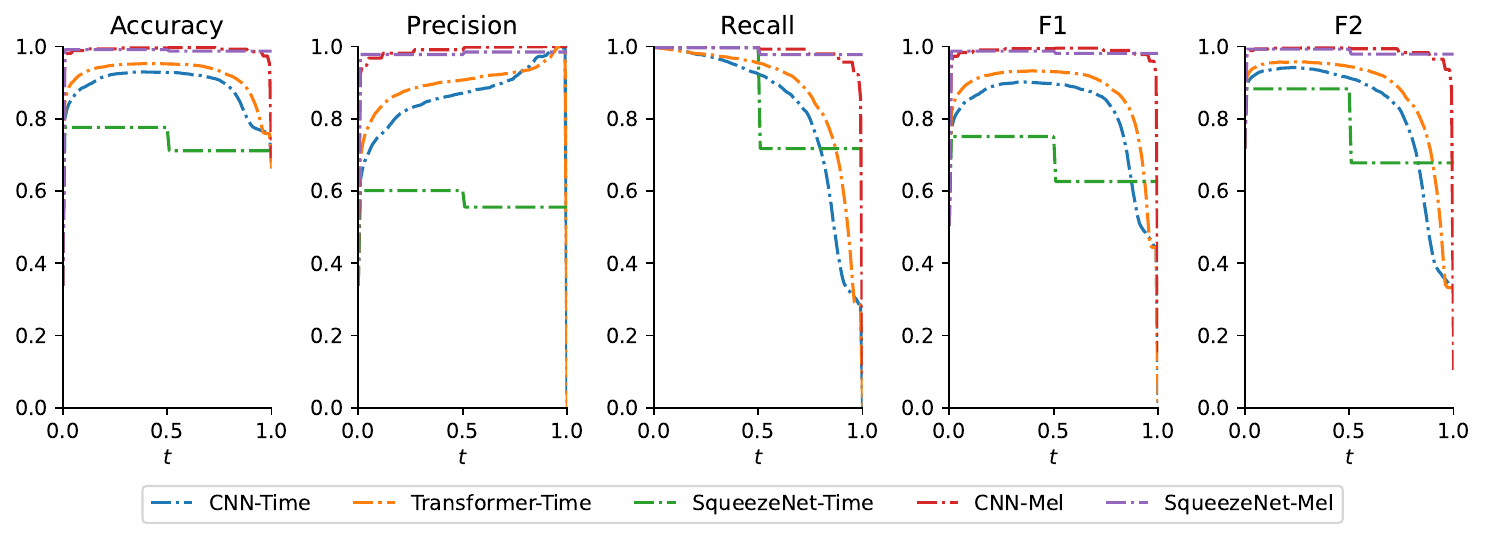}
    \caption{Accuracy, recall, precision, $F$-scores of TinyML models on different threshold values $t$, evaluating on training dataset.}
    \label{fig:prediction_metrics_diff_th_val}
\end{figure*}

We next measured accuracy, precision, recall, $F_1$- and $F_2$-score, shown in Figure~\ref{fig:prediction_metrics_diff_th_val}. Spectrogram-based models achieve the highest metrics for the whole range of threshold values. For time-series models, Transformer-Time worked overall better than CNN-Time; Again, SqueezeNet-Time performs worst (consistent with the above \ac{ROC} analysis). 

Next, we focused on finding the threshold settings for optimal  $F_2$-score prioritizing recall, i.e. reducing the likelihood of mistakenly discarding records of the targeted bird). The results are shown in Table~\ref{tab:mdl_threshold_opt_f2} for training data, thereafter verified on test (previously unseen) data as shown in Table~\ref{tab:eval_test_opt_f2}. Note that this threshold influences energy consumption: more target bird classification leads to more energy consumption (as data needs to we stored on to SD card in this case).

These results are promising and allowing to focus only on the relevant audio segments (10\% of the total recording time), as the classification performance is compelling. Thus, TinyChirp can save 90\% of SD card space compared to the initial monitoring scenario (Section~\ref{subsec:corn_bunt_use_case}), potentially extending the deployment time to 18 weeks. However, complementary experiments must now be carried out on targeted hardware, so as to evaluate other key metrics: computation time, energy consumption and memory footprint on typical microcontroller-based devices. The next sections focus on that part.

\begin{table}[t]
    \centering
    \scriptsize
    \caption{Model threshold and prediction metrics for optimized $F_2$-scores, evaluating on training dataset.}
    \begin{tabular}{llllll}
    \toprule
     Model  & Threshold ($t$) & Acc. & Precision & Recall & $F_2$ \\\midrule
     
     Baseline $t_{low}$ & $1.00 \times 10^{-7}$  & 0.44  & 0.38 & 0.99 & 0.75\\
     Power-saving $t_{high}$ & $1.29 \times 10^{-5}$  & 0.77  & 0.62 & 0.80 & 0.76\\ \midrule
     SqueezeNet-Mel & 0.01 & 0.99 & 0.98 & 1.00 & 0.99\\
     CNN-Mel & 0.27 & 1.00 & 0.99 & 1.00 & 1.00\\
     CNN-Time & 0.23 & 0.92 & 0.82 & 0.98 & 0.94 \\
     Transformer-Time & 0.27 & 0.95 & 0.89 & 0.98 & 0.96\\
     SqueezeNet-Time & 0.01 & 0.78 & 0.60 & 1.00 & 0.88\\

    \bottomrule
    \end{tabular}
    
    \label{tab:mdl_threshold_opt_f2}
\end{table}

\begin{table}[t]
    \centering
    \caption{Evaluation on test dataset using model threshold $t$ for optimized $F_2$-scores.}
    \begin{tabular}{lllll}
    \toprule
     Model  & Accuracy & Precision & Recall & $F_2$ \\\midrule
     Baseline $t_{low}$ & 0.44 & 0.37 & 0.99 & 0.74\\
     Power-saving $t_{high}$ & 0.78 & 0.62 & 0.82 & 0.77\\ \midrule
     SqueezeNet-Mel  & 0.99 & 0.96 & 1.00 & 0.99 \\
     CNN-Mel & 0.99 & 0.98 & 0.99 & 0.99\\
     CNN-Time & 0.94 & 0.86 & 0.98 & 0.95\\
     Transformer-Time & 0.93 & 0.90 & 0.91 & 0.91\\
     SqueezeNet-Time & 0.79 & 0.61 & 1.00 & 0.88 \\

    \bottomrule
    \end{tabular}
    
    \label{tab:eval_test_opt_f2}
\end{table}

\subsection{Performance Evaluation on Microcontrollers}

{\bf Experimental Setup --} For our measurements, we used a common low-power off-the-shelf board: the nRF52840 Development kit (nrf52840dk). This board is based on an ARM Cortex-M4 processor \added{(core frequency at 64 \si{\mega\hertz})}, with a typical memory budget: 1 \si{\mega\byte} of Flash memory and 256 \si{\kilo\byte} of RAM. As embedded software base we used RIOT-ML~\cite{huang2024riot-ml}. 
Throughout the experiments, we monitored the (RAM/Flash) resource footprint using built-in diagnostic tools provided by RIOT-ML and RIOT~\cite{baccelli2018riot}, as well as external measurement equipment as described below.

Energy consumption was measured using an ampermeter to gauge the board's energy efficiency. This involved recording the current draw of the \ac{MCU} and the external storage system (SD card) in different stages. We used a voltage regulator to supply 3.3 \si{\volt} \ac{DC} output and a logger to capture detailed current profiles over time.

{\bf Scope of the Measurements --} Besides the model inference stage, we also considered the resource footprint of pre-processing during bird song recognition. Pre-processing refers to down-sampling and calculation of Mel-spectrogram as described in Section~\ref{subsec:preprocessing}.
Also, the energy consumption in \ac{MCU}'s lowest power mode (idle stage) was measured as the reference of absence of sound with sufficient intensity.

{\bf Measurement Results on Microcontrollers --} The first striking observation is that  SqueezeNet-Time and SqueezeNet-Mel were not deployable due to \ac{OOM} (exceded memory budget) and were thus excluded in this experiment. All other models fit in the constraints of memory and storage budget, even when discounting the memory footprint of the OS and the network stack for wireless communication. 

The resource footprint of different stages and models measured on an nrf52840dk board are shown in Table~\ref{tab:rsrc_cnsump_models}.  We observe that the model inference stage takes the bulk of the overall energy consumption, as it requires the most process time and \ac{MCU} in full power mode. 
As expected, the idle stage requires the least power, as all peripherals and the \ac{MCU} are in the lowest power (standby) mode (until woken up by sound with sufficient amplitude). 

\begin{table*}[h]
    \centering
    \scriptsize
    \caption{Resource consumption of TinyML models on nrf52840dk board.}
    \begin{tabular}{llllllll}
    \toprule
     \multirow{2}{*}{Model / Stage} & \multirow{2}{*}{Memory (\si{\kilo\byte})} & \multirow{2}{*}{Storage (\si{\kilo\byte})} & \multicolumn{2}{c}{Latency (\si{\milli\second})} & \multirow{2}{*}{Power (\si{\milli\watt})} & \multicolumn{2}{c}{Energy (\si{\milli\joule})} \\
                       &                         &                          & Inference          & Preprocess         &           &     Inference      &     Total                                     \\\midrule
     Baseline         & 67.216  & 20.34   & 213.755    & 2.0 & 9.900           & 2.116 & 2.136               \\
     \midrule
CNN-Mel          & 104.328 & 37.868   & 406.146  &  1980.259  & 17.820          & 7.238  & 42.525                 \\
SqueezeNet-Mel    &  OOM  &     -    &   -    &    -   &         -        &   -                      \\\midrule
CNN-Time         & 75.564  & 24.104  & 1490.687  & 2.0 & 17.160          & 25.580 & 25.614                 \\
Transformer-Time & 83.468  & 24.712  & 1079.293  & 2.0 & 17.820          & 19.233 & 19.268                \\
SqueezeNet-Time   & OOM    &         &       &       &                 &                         \\\midrule
Idle       &     -    &   -      &       -       & - &6.270           &     -                    \\
OS with Network Stack    &  7.556       &  35.808       &  -   &  -  &    -        &     -               \\

    \bottomrule
    \end{tabular}
    
    \label{tab:rsrc_cnsump_models}
\end{table*}


{\bf Spectrogram \textit{vs} Time-series --} Considering only the prediction performance in Section~\ref{sec:pred_perf_models}, it would seem natural to choose CNN-Mel as the best candidate, since it also provides the best inference latency. However, taking pre-processing time into account (which for CNN-Mel includes producing a spectrogram) shows a different picture. It takes roughly 2.4 seconds in total for a 3-second audio snippet, which is dangerously near the real-time constraint and costs high power consumption. With the best prediction performance and the lowest total compute latency among time-series models, the Transformer-Time now appears like the best choice to deploy on tiny devices.

Next, we consider jointly the overall resource consumption and the classification performance to discuss variants for the decision strategy described in Section \ref{subsec:decision_strategy}:

{\bf Baseline Only --}
With the \emph{Baseline $t_{low}$} (Step 1 and Step 3 only) setting, the required computational resources are minimal, while achieving a recall of 0.99; however, 
due to the low precision, many false positives are stored, leading to the lowest storage efficiency.

{\bf TinyChirp Skipping Baseline --}
With this variant, the \ac{TinyML} model is applied immediately, without pre-screening by the baseline (Step 2 and Step 3 Only), leading to high predictive performance and high storage efficiency, but requiring higher energy consumption. Even when using the Transformer-Time with the lowest energy footprint, this requires more than 9 times the energy required for the baseline. This variant is ideal for devices currently running out of storage space while still with substantial battery left.

{\bf TinyChirp --}
This variant refers to the full series of Step 1-3 described in Section \ref{subsec:decision_strategy}, whereby the samples not discarded by the baseline are double-checked by the \ac{TinyML} model. This approach offers a good compromise of compute, storage and energy footprint, by discarding early, without impacting too much the overall prediction performance achievable by the \ac{TinyML} model on its own. 

{\bf TinyChirp with Power-saving --}
This variant (Step 1-3 with the power-saving flag enabled) provides a more energy-efficient twist to TinyChirp.
However, the overall precision is determined by the threshold setting \emph{$t_{high}$}, which causes an extra 40\% of false positives to be stored onto the SD card, leading to lower storage efficiency. This variant seems more suitable for currently running out of battery, but with still sufficient storage resources.

\section{Conclusion}
In this paper we present an empirical study of a bioacoustic monitoring use case showing that machine learning is usable directly on autonomous recording units based on low-power microcontrollers. We focused on a concrete use-case: monitoring Corn Bunting birds in rural UK. We publish a data set of the recordings, based on which we demonstrated that TinyML can perform on-device high accuracy classification for bird song classification. Compared to the traditional approach used so far, our approach TinyChirp can extend from 2 weeks to 18 weeks -- almost a full season -- the time intervals until which researchers must harvest overfull data storage embedded on the recording units deployed in the field. Our experiments also observe hands-on how performing inference directly on the audio time-series is the better approach on microcontroller-based hardware, compared to inference on mel-sepctrograms typically used in audio pattern recognition. Looking beyond monitoring Corn Bunting birds' songs, TinyChirp is applicable to a wider range of similar bioacoustic use cases using a large variety of low-power microcontroller-based hardware, based on our general-purpose TinyML approach and our open-source code implementation integrated into the RIOT operating system.

\section*{Acknowledgment}
The research leading to these results partly received funding from the MESRI-BMBF German/French cybersecurity program under grant agreements No. ANR-20-CYAL-0005 and 16KIS1395K. We acknowledge the receipt of media from the Macaulay Library at the Cornell Lab of Ornithology and the Xeno-Canto database.  Felix Bießmann received support from the Einstein Center Digital Future, Berlin,
the German Research Foundation (DFG) (528483508 - FIP 12), the Federal Ministry of Economic Affairs and Climate Action (RIWWER 01MD22007H), the German Federal Ministry of the Environment (GCA 67KI2022A) and the German Federal Ministry
of Education and Research (KIP-SDM 16SV8856).

\bibliographystyle{IEEEtran} 

\bibliography{IEEEabrv,tinyChirp, IEEEsettings}

\end{document}